%% file: main_icra.tex
\documentclass[letterpaper, 10 pt, conference]{ieeeconf}  

\IEEEoverridecommandlockouts                              

\usepackage{times}
\usepackage{epsfig}
\usepackage{graphicx}
\usepackage{amsfonts}
\usepackage{hyperref}
\hypersetup{
    colorlinks=true,
    linkcolor=blue,
    filecolor=magenta,      
    urlcolor=cyan,
}

\usepackage{graphicx}
\usepackage{caption}

\input{shortcuts.tex}
\usepackage{xspace}
\newcommand{\methodName}{\textsc{HandyPriors}\xspace}

\title{HandyPriors: Physically Consistent Perception \\ of Hand-Object Interactions with Differentiable Priors
}

\author{Shutong Zhang$^{1,*}$, Yi-Ling Qiao$^{2,*}$, Guanglei Zhu$^{1,*}$, Eric Heiden$^3$, Dylan Turpin$^{1,3}$, Jingzhou Liu$^{1}$, \\ 
Ming Lin$^{2}$, Miles Macklin$^{3}$, Animesh Garg$^{1,3}$
\thanks{$^{1}$University of Toronto \& Vector Institute, $^{2}$University of Maryland College Park, $^{3}$Nvidia, $^*$ Equal contribution}
\thanks{More experiments and supplementary materials can be found at:
\href{https://handypriors.github.io/}{https://handypriors.github.io/}}
}

\begin{document}

\thispagestyle{empty}
\pagestyle{empty}

\maketitle
\begin{abstract}
Various heuristic objectives for modeling hand-object interaction have been proposed in past work. However, due to the lack of a cohesive framework, these objectives often possess a narrow scope of applicability and are limited by their efficiency or accuracy. In this paper, we propose \methodName, a unified and general pipeline for pose estimation in human-object interaction scenes by leveraging recent advances in differentiable physics and rendering. Our approach employs rendering priors to align with input images and segmentation masks along with physics priors to mitigate penetration and relative-sliding across frames. Furthermore, we present two alternatives for hand and object pose estimation. The optimization-based pose estimation achieves higher accuracy, while the filtering-based tracking, which utilizes the differentiable priors as dynamics and observation models, executes faster. We demonstrate that \methodName attains comparable or superior results in the pose estimation task, and that the differentiable physics module can predict contact information for pose refinement. We also show that our approach generalizes to perception tasks, including robotic hand manipulation and human-object pose estimation in the wild.
\end{abstract}

\section{Introduction}

\input{tex/introduction.tex}

\section{Related Works}
\input{tex/related-works.tex}

\section{Methods: \methodName}
\input{tex/methods.tex}

\section{Experiments}
\input{tex/experiments.tex}

\section{Conclusion}
\input{tex/conclusion}

\clearpage
{\small
\bibliographystyle{IEEETran}
\bibliography{refs}
}

\end{document}

%% file: tex/introduction.tex
Hands serve as the primary means by which humans interact with the physical world as they are responsible for most of the intricate and dexterous object manipulation tasks.
A good understanding of hand-object interaction helps with numerous learning-based tasks such as action recognition~\cite{garcia2018first},
robotic manipulation~\cite{andrychowicz2020learning},
user interaction~\cite{sharma2015human}, etc.  

Estimating hand and object interaction presents several challenges.
Firstly, the complex, deformable geometry and dynamics of hands present challenges in learning and optimization.
Unlike rigid objects, hands are articulated with joints and possess high degrees of freedom.
Their non-linearity and non-convexity further exacerbate the complexity of the task.
Secondly, the presence of both self-occlusion and hand-object occlusion result in uncertainty when predicting hand poses during manipulation.
Finally, the rich contacts between hands and objects in interaction scenes necessitate the need for a consistent estimation method for both the hand and the object to avoid penetration.

There are generally two categories of approaches for estimating hand and object poses. The first approach is purely learning-based, where the network architectures, loss functions, and training strategies are carefully designed to train a generalizable mapping from images to hand and object configurations~\cite{hasson19_obman,liu2021semi}.
Although neural networks and large training datasets can mitigate the difficulties posed by high degrees of freedom and non-convexity, they cannot guarantee interpretability nor correctness. The second category of approaches is optimization-based~\cite{rhoi2020,hasson20_handobjectconsist},
which involves defining and optimizing heuristic target functions to improve physical feasibility and reduce penetrations. Although the optimization process may be slower, per-frame iterations can improve the accuracy of the results. 

\begin{figure}
    \centering
    \includegraphics[width=0.95\linewidth]{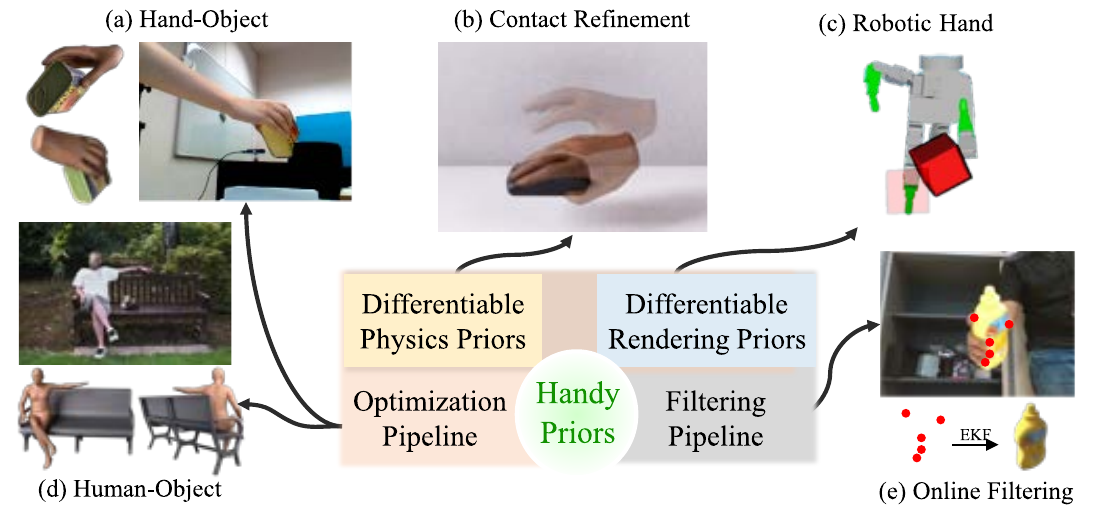}
    \vspace*{-1em}
    \caption{\textbf{\methodName is a modular block application in multiple tasks.} The estimation of hand-object interaction can be achieved by (a) optimization-based refinement or (e) filtering with Extended Kalman Filter (EKF). Moreover, the differentiable contact module can be used to perform (b) pose refinement given contact status. Our differentiable priors can also be used in pose estimation for scenes with (c) robot hands or (d) human bodies. }
    \label{fig:teaser}
    \vspace*{-2em}
\end{figure}

In this paper, we leverage both learning and optimization techniques. Our method utilizes pre-trained networks to provide an initial estimate of poses, and then uses differentiable self-supervised terms to further refine the estimation by correcting errors from pixel-level misalignment and non-physical penetration/oscillation. Inspired by the success of self-supervised learning in large-scale vision and language models~\cite{he2022masked,brown2020language}, we adopt similar techniques to estimate hand-object interactions. Specifically, we propose {\em supervising the pose prediction by comparing it with subsequent video frames}, similar to the way in which text generation can be trained using sequence completion.

To enable backpropagation of gradients from image space, we develop {\em fully-differentiable operators that simulate and render predicted poses}. This closed-loop optimization approach improves the accuracy by making the estimation resemble more of the input image. We also {\em incorporate several regularization terms to minimize motion oscillation and collision in predicted sequences}. 

While the optimization process improves the accuracy, it can be time-consuming. To address this issue, we present an alternative filtering technique based on the Kalman filter~\cite{ribeiro2004kalman} to further {\em utilize our differentiable dynamics and observation modules}. This approach allows the user to make a trade-off between efficiency and accuracy depending on whether they choose to perform online tracking or offline optimization. Employing this integrated differentiable rendering and simulation pipeline, we also apply our pose estimation method to human and synthetic robot hand scenes, demonstrating the generalizability and robustness of our method. Figure~\ref{fig:teaser} summarizes several applications of this approach.

\vspace*{0.25em}
In summary, the key contributions of this work are:
\begin{itemize}
\setlength{\itemsep}{1mm}
\setlength{\parskip}{0pt}
    \item An {\bf integrated differentiable rendering} (Sec.~\ref{sec:rendering}) and {\bf simulation} (Sec.~\ref{sec:physics}) pipeline to estimate hand-object interaction. 
    \item An {\bf offline optimization} (Sec.~\ref{sec:optimization}) and an {\bf online tracking} (Sec.~\ref{sec:tracking}) process that can predict poses and contact information, providing the user with options to balance between accuracy and efficiency. 
    \item {\bf Experiments that generalize our priors to applications}, e.g. robotic manipulation (Sec.~\ref{sec:exp_robot}), contact refinement (Sec.~\ref{sec:exp_contact}),  and human bodies (Sec.~\ref{sec:exp_human}).
\end{itemize}

In our experiments, our method achieves comparable or better pose estimation results in hand-object-interaction datasets~\cite{hampali2020honnotate,hampali2022keypointtransformer}. Additionally, we show that the contact information predicted by our method effectively refines pose estimation. Our ablation study compares the differences between the optimization and tracking results, further validating the contributions of different losses.

%% file: tex/related-works.tex
\begin{figure*}
    \centering
    \includegraphics[width=1\linewidth]{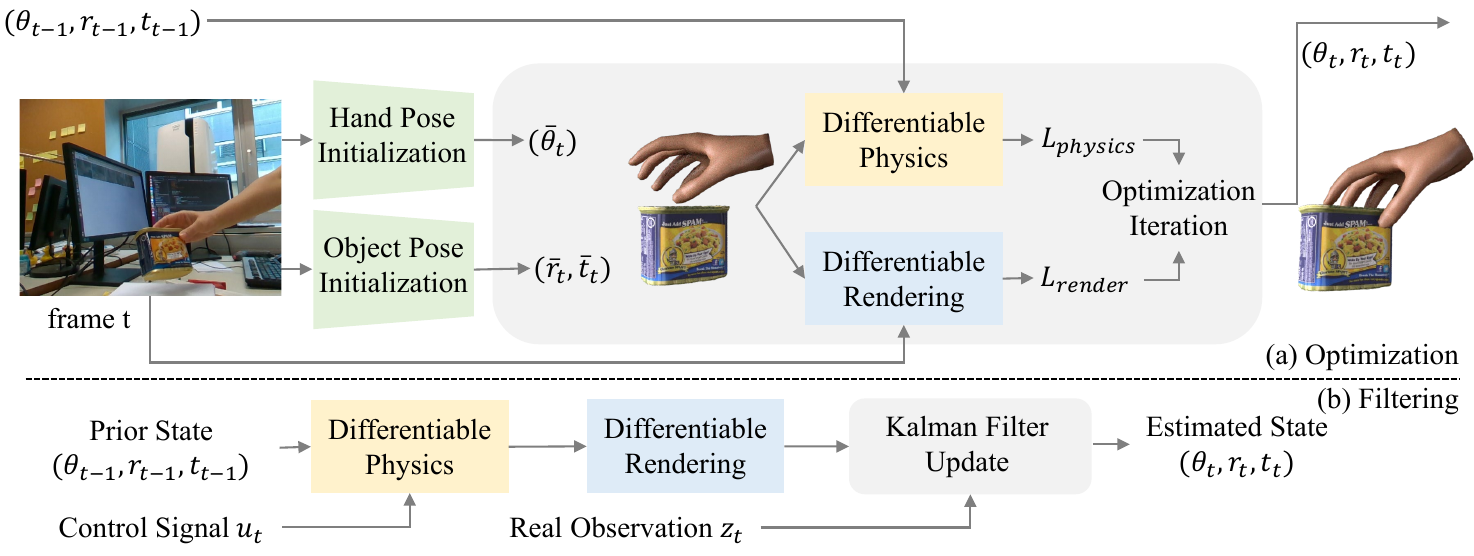}
    \vspace{-2em}
    \caption{\textbf{An overview of our optimization and filtering pipelines.} We provide two alternatives for utilizing the differentiable priors. Given the image and the estimation from previous frames, (a) the optimization pipeline first initializes the poses with pre-trained networks and then minimizes the rendering and physics losses from the differentiable operators; (b) the filtering pipeline can take some simple observations and use Extended Kalman Filter (EKF) to update the state estimation. EKF requires differentiable physics and rendering to model the system and runs much faster than the optimization pipeline. }
    \label{fig:overview}
    \vspace{-2em}
\end{figure*}



\mypara{Hand and object pose estimation} from monocular images~\cite{billings2019silhonet,Eric14,hu2019segpose,tekin18} is a fundamental problem in computer vision. Many recent works predict 6 DoF object poses assuming known rigid mesh models~\cite{He_2021_CVPR,wang2019densefusion,xiang2018posecnn}. 
For human hands, MANO~\cite{MANO:SIGGRAPHASIA:2017} proposes a parametric model to represent the hand geometry using different shape and pose parameters. Numerous learning-based approaches focus on predicting 3D joint locations from RGB(D) inputs~\cite{Cai2018WeaklySupervised3H,Ge_2018_HandPointNet,GANeratedHands_CVPR2018,spurr2018cvpr,Christian_2017_ICCV,Yang_2019_CVPR,Yuan_2018_CVPR}. Others~\cite{Beak_2019_cvpr,boukhayma20193d,hasson20_handobjectconsist,rong2021frankmocap} then regress to the MANO pose and shape parameters. 
There are also approaches directly learning a mapping between visual inputs and hand poses from annotated datasets~\cite{Freihand2019,Kulon_2020_CVPR,Moon_2020_ECCV_InterHand2.6M,OccludedHands_ICCV2017}. 
These methods achieved remarkable success on hand-only datasets but had poor performance when hands were holding objects. 
There are also optimization-based approaches that optimize hand parameters based on 2D key points~\cite{rhoi2020,hasson20_handobjectconsist,Panteleris_2018_WACV} or segmentations~\cite{hampali2020honnotate}. They tend to have better accuracy but are much slower due to the iterative optimization process. We propose a tracking-based pipeline with higher efficiency. Our differentiable priors are {\em more general} and can be applied to a {\em wider range} of applications.

\mypara{Hand object interaction} has been garnering progressively more attention. Treating hand and object pose estimation separately results in inaccurate estimation for both. With an increasing number of hand-object datasets being created~\cite{Liu_2022_CVPR,Brahmbhatt_2020_ECCV,chao:cvpr2021,hampali2020honnotate,hasson19_obman,FirstPersonAction_CVPR2018}, many recent works have started to focus on jointly tracking hands and objects. However, simultaneously estimating both is challenging due to occlusions and depth ambiguity. \cite{liu2021semi} designed a joint learning framework based on contextual reasoning between hand and object representations. \cite{yang2021ArtiBoost} proposed an ordinal relation loss to correct depth misalignment between hand and object. \cite{Park_2022_CVPR_HandOccNet} incorporated a feature injection mechanism that better handles occlusion by integrating hand information into the obscured areas. These methods improve the accuracy of pose estimation but pay little attention to the physical interaction between hand and objects.
Other works aim to predict physically plausible hand-object poses. To achieve this goal, interaction constraints are applied to encourage contact and minimize interpenetration. \cite{hasson20_handobjectconsist,rhoi2020} proposed to use signed distance functions (SDF) to model contact between hand and object, ~\cite{grady2021contactopt} design a deep network to estimate contact areas and a virtual capsule technique to simulate soft hand tissue deformation. 
\cite{Haoyu_2022_sa} apply motion and force vectors to reconstruct interactions. Compared to these methods, our approach is more general and can be used in {\em contact-based refinement and pose estimation in both human-hand and non-human-hand scenarios}.

\mypara{Differentiable Priors.}
Differentiable rendering~\cite{liu2019softras,Laine2020,david2019mitsuba} and physics~\cite{heiden2020NeuralSim,xu2021end, heiden2022pds} have been used to solve inverse problems for their efficiency in optimization.  
For hand grasping, \cite{christen2022dgrasp} leverages a physics simulator to evaluate dynamic interactions between the hand and the object. ~\cite{Dylan_eccv_2022}, \cite{turpin2023fastgraspd} uses differentiable physics to generate physically realistic and stable grasps.
There are also recent works that use differentiable rendering to optimize human~\cite{dwivedi2021learning} or object~\cite{zhang2020phosa} poses. However, they do not consider their physical interactions. \cite{gartner2022differentiable} utilize a differentiable articulated body simulator to learn and smoothen human-pose estimation. In contrast, our method leverages rendering priors for each frame, thereby significantly accelerating pose estimation in a video.

%% file: tex/methods.tex
\subsection{Problem Setting}
Given a sequence of RGB hand-object-interaction images and the object's 3D model, we propose two pipelines to estimate the configuration of the hand and object in each frame. Figure~\ref{fig:overview} shows an overview of our method.

In particular, we represent the hand using the MANO~\cite{MANO:SIGGRAPHASIA:2017} parameterization, which describes a hand by its pose $\theta\in \mathbb{R}^{45+6}$ (representing the rotations of 15 hand joints and a 6 DoF global transformation) and shape $\beta\in \mathbb{R}^{10}$ (a statistical model) parameters. The 778 vertices $\VV_{hand}\in\mathbb{R}^{778\times 3}$ of the hand mesh can be computed from $\theta, \beta$ using a learned linear model $\VV_{hand}=m(\theta, \beta)$. 

The object pose has 6 degrees of freedom, which contains its translation $\tt\in R^{3}$ and rotation (represented by XYZ Euler angles) $\rr\in R^{3}$. 

%
%

\subsection{Differentiable Rendering Priors} 
\label{sec:rendering}
The first prior of our approach is differentiable rendering, which helps align the estimated 3D model with the 2D image.
Instance segmentation is the key information we rely on to optimize hand and object poses. We use a segmentation detector~\cite{Kirillov_2020_pointrend} trained on the COCO dataset~\cite{Lin_coco_2014} to extract hand and object masks. Our experiment shows that although the image segmentation detector, {\em PointRend}~\cite{Kirillov_2020_pointrend}, was not trained on the YCB objects ~\cite{Berk_2015_YCB}, it can correctly predict object masks. However, it fails to adequately segment hands as fingertips are harder to detect. Thus, we further trained {\em PointRend} to obtain better segmentations for the hand and hand-held objects.

For each frame, we utilize a differentiable renderer {\em SoftRas}~\cite{liu2019softras} that rasterizes and projects 3D hand and object models to 2D, then we optimize hand parameters $\theta$ and object parameters $\rr, \tt$ by fitting the projected vertices onto the hand and object masks.
Next, we describe loss terms related to the rendering part.

\mypara{RGB image loss.} Since hand texture varies greatly from person to person, we only render the RGB images for objects. We then apply L2 loss between the rendered image $I_{r}$ and the input RGB image $I_{in}$ cropped by the detected object mask $M_{obj}$. 
The image loss term can be expressed as follow:
\begin{equation}
    L_{img}(\rr,\tt) = \lVert (1 - M_{hand})I_r - M_{obj}I_{in} \rVert_{2}^{2}
\end{equation}

\mypara{Mask loss.} Compared to the RGB image loss, masks have fewer features but are not sensitive to lighting and texture, hence the loss based on masks is more robust and general. We use a differentiable renderer to render the silhouette of hands and objects together. We then apply L2 loss between the rendered mask  $M_{r}$ and the detected mask $M_{d}$,
%
\begin{equation}
     L_{mask}(\theta, \rr,\tt) = \lVert M_{r} - M_{d} \rVert_{2}^{2}
\end{equation}
%

\subsection{Differentiable Physics Priors}
\label{sec:physics}
Besides the rendering priors defined on the 2D image spaces, we also add more physics-related priors in the 3D and temporal domain to help us regularize the hand-object understanding.

\mypara{Contact loss}. During manipulation, we observe that most of the contacting vertices between hand and objects do not have sudden relative sliding. Based on this observation, we first utilize a GPU-based differentiable contact module to infer the signed distance function $sdf(\VV_{hand})$ from hand vertices to the object. Then, we find the contact points $\vv_{t}$ where the SDF values are smaller than 0.1. The relative sliding between $t$ and $t-1$ is,
$$
L_{sliding} (\theta, \rr, \tt) =\lVert (\vv_{t}-\rr_t)(\rr_{t}^{-1})\trans - (\vv_{t-1}-\tt_{t-1})(\rr_{t-1}^{-1})\trans \rVert
$$

Moreover, given the SDF results, we reduce penetration by penalizing the negative SDF values.
\begin{equation}
L_{penetration}(\theta, \rr, \tt) = \min(0, -sdf(\VV_{hand}))
\end{equation}

\mypara{Smoothness loss.} To enhance the continuity of hand and object across time, we penalize sudden changes of object and hand in two consecutive frames. For objects, we apply L2 loss between object vertices in the current frame $\VV_t$ and previous frame $\VV_{t-1}$. For hands, we compute the L2 loss between hand poses in two frames $\theta_t, \theta_{t-1}$, 
\begin{equation}
     L_{con} = \lVert \VV_t - \VV_{t-1} \rVert_{2}^{2} + \lVert \theta_t - \theta_{t-1} \rVert_{2}^{2}.
\end{equation}

 During the experiment, we also observed that objects usually have jittering for several frames. Thus, we add a smoothness term to punish the oscillation in object velocity, 
\begin{equation}
     L_{smo} = \lVert (\VV_t - \VV_{t-1}) - (\VV_{t-1} - \VV_{t-2}) \rVert_{2}^{2}.
\end{equation}


\subsection{Optimization-based Refinement}
\label{sec:optimization}
In summary, the losses of rendering priors are 
\begin{equation}
L_{render}=\lambda_1L_{image}+\lambda_2L_{mask},
\end{equation}
and the physics-inspired losses are 
$$
L_{physics}=\lambda_3L_{sliding}+\lambda_4L_{penetration}+\lambda_5L_{con}+ \lambda_6L_{smo},
$$
\noindent
where $\lambda$'s are the weights for the loss terms.
With these differentiable loss terms, we optimize the pose of the hand $\theta$ and object $\rr,\tt$ using a gradient-based optimization method such as Adam~\cite{Adam2014}.
\begin{figure}
    \centering
    \includegraphics[width=0.8\linewidth]{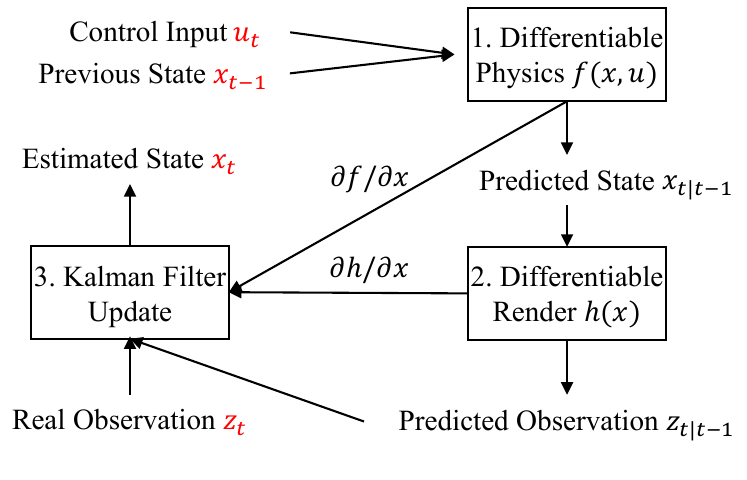}
    \vspace*{-1.5em}
    \caption{\textbf{Diagram of the EKF filtering process.} With (1) the differentiable physics $f$ and (2) rendering $h$ modules to compute the Jacobians, (3) the extended Kalman Filter (EKF) can perform state estimation given the control signal, observation, and previously estimated states.  }
    \label{fig:filter}
\end{figure}

\begin{table}[]
\caption{\textbf{Ablation study.} We run ablation studies for both the optimization (left column) and filtering pipeline (right column). For optimization, 2D and 3D errors ($cm$) decrease with gradually added rendering $\mathcal{L}_{render}$ and physics loss $\mathcal{L}_{physics}$. Filtering is much faster than the optimization pipeline but also scarifies accuracy.}
\vspace{-0.5em}
\resizebox{1\linewidth}{!}{
\begin{tabular}{lccc|lccc}
Optimization  & 2D Error & 3D Error & Runtime (s)   & Filtering & 2D Error & 3D Error & Runtime (s) \\
\midrule
Initialization  & 1.03 & 52.5 &  0.072& $\mathcal{O}_{hand}$ & 44.59 & \textbf{5.57} & \textbf{0.024}\\
+$\mathcal{L}_{render}$     & 0.77  & 1.54 & 2.29& $\mathcal{O}_{object}$ & 19.00 & 11.24 & 0.027\\
+$\mathcal{L}_{physics}$  & \textbf{0.43} & \textbf{1.10} & 5.83& $\mathcal{O}_{hand+object}$ & \textbf{8.84}  & 7.41 & 0.029  \\
\bottomrule
\end{tabular}
}
\vspace{-1em}
\label{tab:ablation}
\end{table}

\subsection{Filtering-based Tracking}
\label{sec:tracking}
Besides the optimization-based process, we can also use the differentiable priors in a more light-weight pipeline (Fig.~\ref{fig:filter}). Multiple filtering algorithms can be integrated with this method for fast tracking; in this work, we employ the Extended Kalman Filter (EKF)~\cite{ribeiro2004kalman} as an example. EKF can give a state estimation in a dynamic system
\begin{equation}
    \xx_t=f(\xx_{t-1}, \uu_t) + \epsilon_{s},
    \zz_t=h(\xx_t)+\epsilon_{o},
\end{equation}
where $\xx$ is the state, $\uu$ is the control, $\zz$ is the observation, $\epsilon_{s}$ and $\epsilon_{o}$ are noises, $f(\cdot),h(\cdot)$ are \textit{differentiable} functions for states and observation, respectively. Using the EKF, we can compute an estimation $\hat{\xx}_t$ for the hidden state $\xx_t$ given the observation $\zz_t$ and control $\uu_t$.

The states of the hand-object system are $\xx_t=(\theta_t, \rr_t, \tt_t)$. Using those pose parameters (along with the known object model and shape parameters $\beta$), we can reconstruct the entire scene. For most of the cases in real life and nearly all of the cases in existing datasets, hands can be modelled as actuators whereas objects are primarily passive. Therefore, the control input for the system is the hand pose $\uu_t=\theta$. 

Since the differentiable renderer can output various quantities (2D keypoints, bounding box, full image set), the observations we can choose are also flexible. For instance, the 2D positions of the fingertips can serve as the only feedback supervision. Then the observation function is, 
\begin{equation}
\label{eq:observation}
    h(\xx_t) = S_{tips}\cdot\VV_{hand}(\theta_t),
\end{equation}
where $S_{tips}\in\mathbb{R}^{5\times 778}$ is a one-hot encoding matrix that selects fingertips out of the hand.

The more challenging part is to construct the dynamics function $f(\cdot)$. A direct solution is to use a differentiable full dynamics model to describe the system. However, the actuation signals of the hand (like muscle contraction) are unavailable from the MANO parameter. We thus choose to use simpler and more robust quasistatic dynamics. If the hand makes contact with an object on vertices $\vv_{t-1}=S_{contact}\cdot\VV_{hand}(\theta_{t-1})$ at time step $t-1$, we assume the relative sliding on the vertices from $\vv_{t-1}$ to $\vv_{t}\in\mathbb{R}^{n_{contact}\times 3}$ is small. The rotation $\RR$ and translation $\TT$ are estimated as,
\begin{align}
\label{eq:dynamics}
    \UU\Sigma\VV^\dagger&=svd(\Tilde{\vv}_{t-1}\trans\Tilde{\vv}_t)\\ \nonumber
    \RR_{t} &= (\UU\VV^\dagger)\trans \RR_{t-1}\\ \nonumber
    \TT_{t} &= \Bar{\vv}_t - \Bar{\vv}_{t-1}(\UU\VV^\dagger) + \TT_{t-1}
\end{align}
where $\Tilde{\vv}=\vv-\Bar{\vv}$ is the contact vertices subtracting their centroids $\Bar{\vv}$, $svd(\cdot)$ is the singular value decomposition. 

With the differentiable observation $h(\cdot)$ (Eq.~\ref{eq:observation}) and dynamics $f(\cdot)$ (Eq.~\ref{eq:dynamics}), we can then use the EKF to track and estimate the states $\xx_t=(\theta_t, \rr_t, \tt_t)$.


%% file: tex/experiments.tex
In this section, we apply our differentiable priors to five scenarios: 
(a) object pose estimation for robotic manipulation,
(b) pose estimation, (c) filtering-based tracking, (d) contact refinement,  and (e) human-object pose estimation. Our method can achieve comparable or 
better results in a wide range of tasks.

\begin{figure}
\centering
\begin{tabular}{@{}c@{\hspace{0.3mm}}c@{\hspace{0.3mm}}c@{}}
    \includegraphics[width=0.32\linewidth]{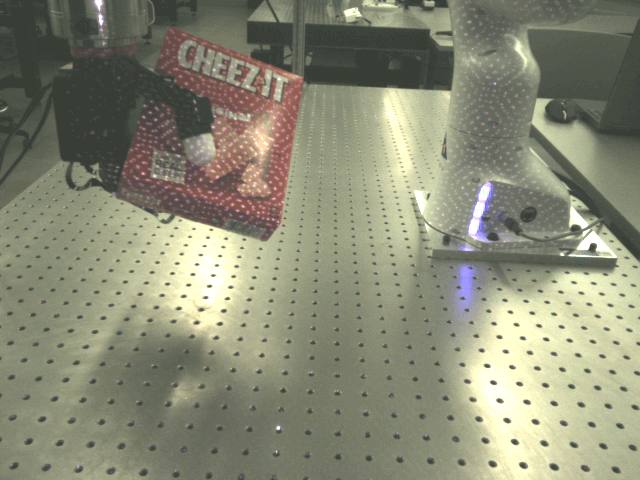}&
    \includegraphics[width=0.32\linewidth]{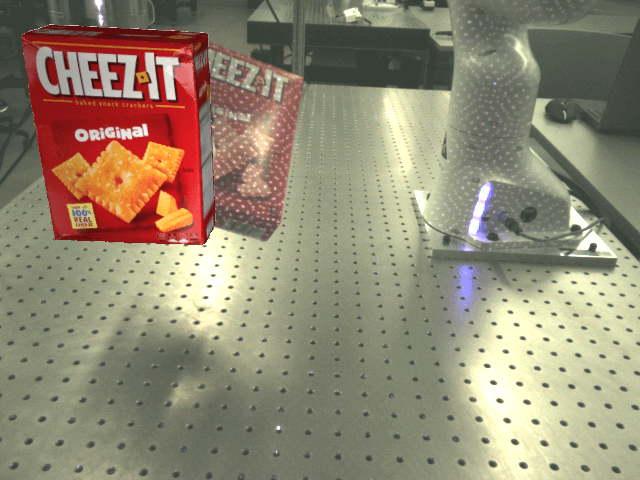} &
    \includegraphics[width=0.32\linewidth]{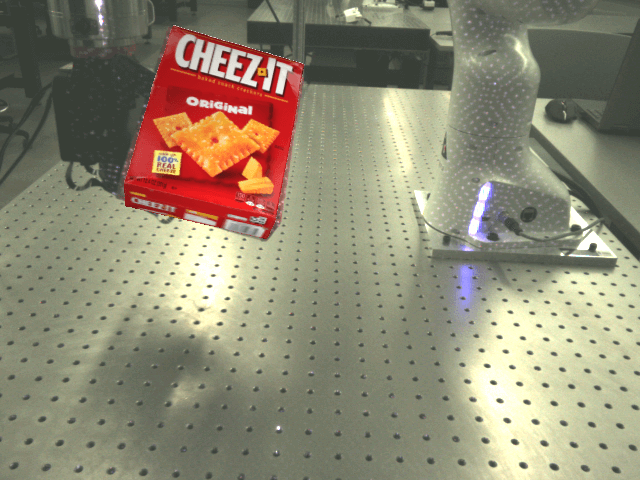} \\
    \small (a) Input & \small (b) Semi~\cite{liu2021semi}  &  \small (c) \methodName \\
\end{tabular}
\vspace{-2mm}
\caption{\textbf{Real-world pose estimation.} (a) We record a robotic hand manipulating a cracker box in the real world using a calibrated camera. With (b) initialization using~\cite{liu2021semi}. We observe that (c) our method can recover the transformation of the box.}
\label{fig:robotic}
\vspace{-2em}
\end{figure}

\subsection{Real-world Robotic Hand and Object Iteration}
\label{sec:exp_robot}
Our method can generalize to a wide range of scenarios, including robotic hands and object interaction. To demonstrate this, we designed a robotic hand manipulation task.
We first record sequences of RGB images where an Allegro robotic hand interacts with a YCB object~\cite{Berk_2015_YCB} using the calibrated Intel RealSense camera and record the corresponding hand pose parameters.
We then predict the initial object pose using only the RGB image with ~\cite{liu2021semi}.
Given the robotic hand configuration calculated using the hand pose parameter and segmentation masks predicted by Pointrend~\cite{Kirillov_2020_pointrend}. We then use our pipeline to optimize the object pose $\rr, \tt$. Figure~\ref{fig:robotic} shows the optimization result of our method. The results demonstrate that our method is capable of generating accurate object poses even with robotic hand interaction scenes.

\subsection{Pose Estimation}
\begin{figure*}
\centering
\begin{tabular}{@{}c@{\hspace{1mm}}c@{}}
    \raisebox{0.5mm}{\rot{HandyPriors}} & 
    \includegraphics[width=0.95\linewidth]{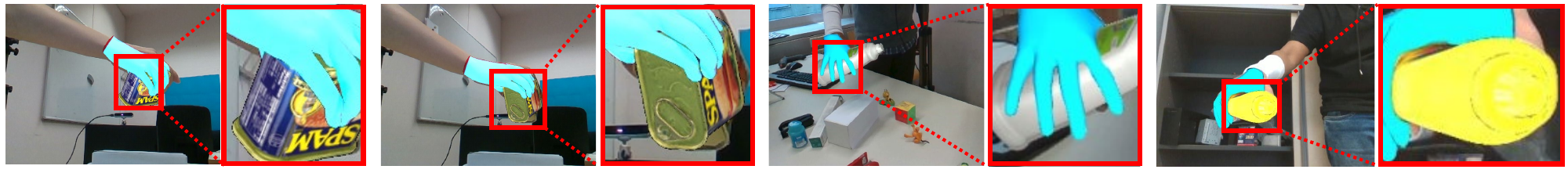}  \\
    \raisebox{3mm}{\rot{Semi~\cite{liu2021semi}}} & 
    \includegraphics[width=0.95\linewidth]{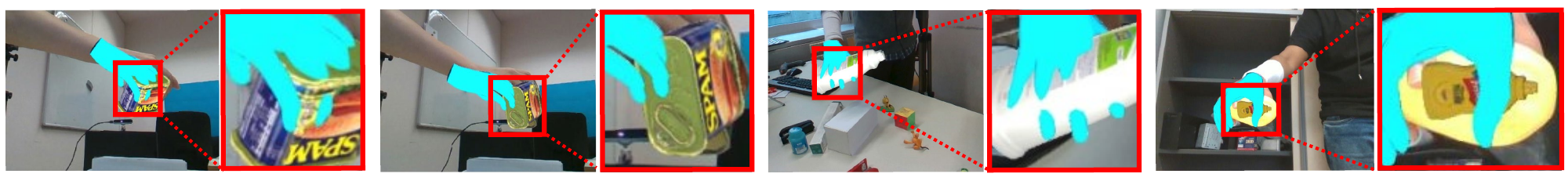} 
\end{tabular}
\vspace{-4.5mm}
\caption{\textbf{Pose Estimation Results.} We compare the hand and object pose estimation results with other methods and visualize the 3D meshes projected on the original image. Our method yields reasonable joint estimation for the interaction scenes.}
\vspace{-1em}
\label{fig:pose_estimation}
\end{figure*}

\begin{table}[]
\caption{\textbf{Evaluation of hand pose estimation.} We measure the joint error ($cm$), mesh error ($cm$), and joint AUC on the DexYCB dataset~\cite{chao:cvpr2021}. PA is Procrustes Alignment. 
}
\vspace{-0.5em}
\resizebox{1\linewidth}{!}{
\begin{tabular}{llcccc}
Methods & Joint (PA) & Joint & Mesh (PA) & Mesh& Joints AUC  \\
\midrule
Artiboost~\cite{yang2021ArtiBoost} & 0.65 & 1.12  & 0.69  & 1.21  &  0.870 \\
Homan~\cite{hasson20_handobjectconsist}    & 1.84  & 4.41& 1.80  & 4.34 &  0.644 \\
HandOccNet~\cite{Park_2022_CVPR_HandOccNet} & 0.62 & 1.16 & 0.66 & 1.20 &  0.877 \\
\methodName (ours) & \textbf{0.52} & \textbf{0.81} 	& \textbf{0.51} & \textbf{0.89} &  \textbf{0.896} \\
\bottomrule
\end{tabular}
}
\vspace{-4mm}
\label{tab:hand}
\end{table}

\subsubsection{Datasets and Metrics}
\mypara{Datasets}. 
HO3D dataset~\cite{hampali2020honnotate} is a real hand-object interaction dataset that contains approximately 80K images from more than 65 sequences. 
It uses 10 models from the YCB dataset~\cite{Berk_2015_YCB}. 
The evaluation of the HO3D dataset is done via an online server.
We report the performance of all of the methods on the evaluation split of the dataset.
Dex-YCB dataset~\cite{chao:cvpr2021} consists of 582K images over 1,000 sequences of 10 subjects grasping 20 different YCB objects~\cite{Berk_2015_YCB} from 8 camera views.
It is currently the largest real hand-object interaction dataset.
In this work, we report the performance of our method on the $S_0$ test split of the dataset.
Following~\cite{yang2021ArtiBoost}, we filter out frames that contain the left hand or where the hand is not in contact with the object. 
\mypara{Metrics}. 
In Table~\ref{tab:pose_comparison}, we measure the 2D error by projecting the object vertices onto the image plane and computing the Chamfer distance between the ground truth vertices and predicted vertices. The 3D error is the Chamfer distance in the original 3D space.
We also report the average interpenetration volume between the hand and the object to measure how well the methods handle interactions. This metric is denoted as Collision. In Table~\ref{tab:hand}, we compute the joint error, mesh error, and joint AUC (Area Under the ROC Curve). For joint and mesh error, we report results before and after Procrustes alignment, where Procrustes alignment changes the global rotation, translation, and scale of the estimated pose to match the ground truth.

\begin{table}[]
\caption{{\bf Quantitative evaluation of object pose estimation}. Our approach is compared with Semi~\cite{liu2021semi}, Artiboost~\cite{yang2021ArtiBoost}, Homan~\cite{hasson20_handobjectconsist} on the HO3D~\cite{hampali2020honnotate} and DexYCB Dataset~\cite{chao:cvpr2021}. We report the 2D ($cm$) and 3D ($cm$) errors. 
}
\vspace{-0.5em}
\resizebox{1\linewidth}{!}{
\begin{tabular}{@{}lccccc@{}}
  \toprule
  Method & 2D Error (HO3D) &  3D Error (HO3D) &  2D Error (DexYCB) &  3D Error (DexYCB)  \\
  \midrule
  Artiboost~\cite{yang2021ArtiBoost}  & 6.41  & 10.83 & 3.01 & 6.78 \\
  Homan~\cite{hasson20_handobjectconsist}   & 4.60 & 41.33 & 15.21  & 95.07     \\
  Semi~\cite{liu2021semi}   & 2.06 &26.35 & 9.20 & 47.77  \\
  \methodName (ours)  & \textbf{1.67} & \textbf{3.22} & \textbf{0.96} & \textbf{3.79} \\
  \bottomrule
\end{tabular}
}
\label{tab:pose_comparison}
\vspace*{-2em}
\end{table}

\subsubsection{Optimization-based Refinement}
We compare our optimization-based method described in Section~\ref{sec:optimization} with Semi~\cite{liu2021semi}, Artiboost~\cite{yang2021ArtiBoost}, Homan~\cite{hasson20_handobjectconsist} for the pose estimation task. In the HO3D dataset, we test three objects from the evaluation split: Mustard Bottle (SM11), Meat Can (MPM10-15), and Bleach Cleanser (SB11, 13). Table~\ref{tab:pose_comparison} shows that our method outperforms alternative methods for estimating the pose of the objects. Our method also achieves comparable results in terms of collision metrics. Table~\ref{tab:hand} shows the hand prediction results compared to Artiboost~\cite{yang2021ArtiBoost}, Homan~\cite{hasson20_handobjectconsist}, and HandOccNet~\cite{Park_2022_CVPR_HandOccNet}. Our method also improves over the initialization from \cite{Park_2022_CVPR_HandOccNet} and achieves the best accuracy in all metrics. Figure~\ref{fig:pose_estimation} shows some qualitative results of the predictions. Compared to other methods, our optimized results are better aligned with the image and have more reasonable relative positions between hands and objects.

\subsubsection{Ablation Study}
In this part, we study the influence of different loss terms on the estimation results. All the experiments are run in the evaluation set SM1 from the HO3D dataset~\cite{hampali2020honnotate}. Table~\ref{tab:ablation} presents the quantitative results. Starting from the initialization by the pre-trained network~\cite{liu2021semi}, adding rendering loss $\mathcal{L}_{rendering}$ drastically improves the performance, especially with regards to the 3D error as it can adjust the translation errors in the depth direction. Moreover, adding the physics loss term $\mathcal{L}_{physics}$ improves the results further by reducing penetration and oscillation in the predicted sequences.

\begin{figure}
\centering
\begin{tabular}{@{}c@{\hspace{0.3mm}}c@{\hspace{0.3mm}}c@{\hspace{0.3mm}}c@{}}
    \includegraphics[width=0.25\linewidth]{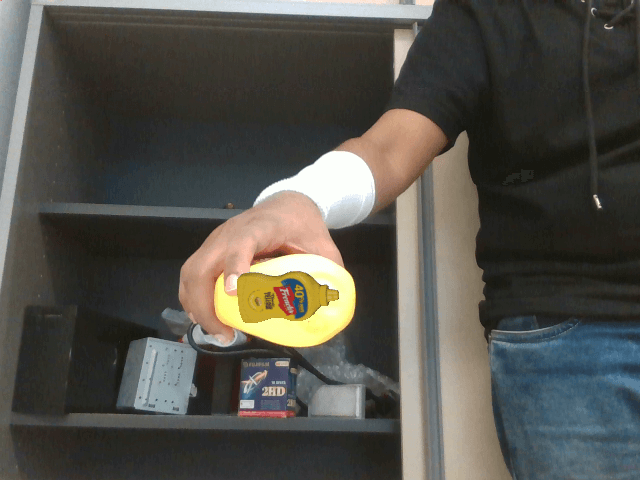} &
    \includegraphics[width=0.25\linewidth]{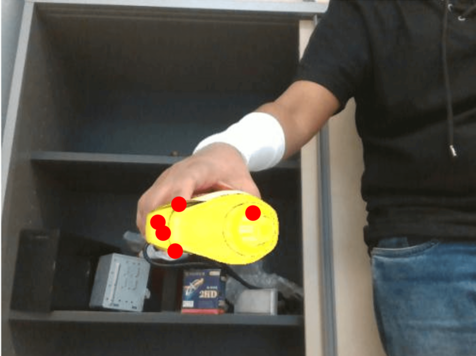}  &
    \includegraphics[width=0.25\linewidth]{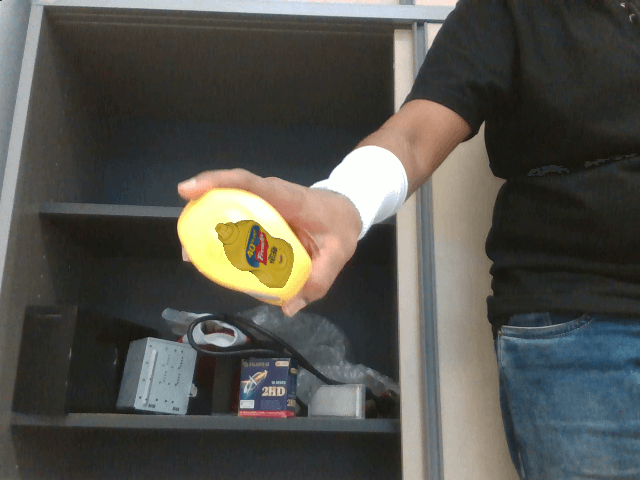} &
    \includegraphics[width=0.25\linewidth]{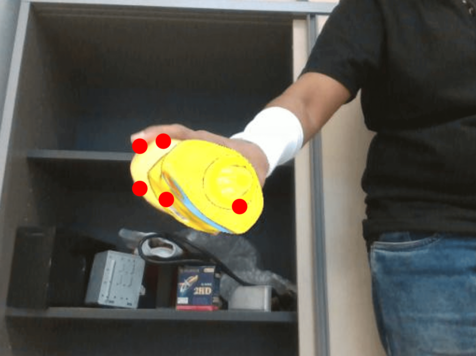}  \\
    \small (a) Semi~\cite{liu2021semi} & \small (b) Filtering &  \small (c) Semi~\cite{liu2021semi} & \small (d) Filtering  \\
\end{tabular}
\vspace{-2mm}
\caption{\textbf{Filtering results.} We show that the filtering pipeline can estimate the object pose with just fingertips (red points) as input. It can give robust and continuous estimation even when the purely learning-based method (a, c) fails. }
\label{fig:comp_simulation}
\end{figure}

\subsection{Filtering-based tracking}
\label{sec:exp_filter}
We apply the tracking algorithm described in Section~\ref{sec:tracking} to perform pose estimation. We run on the SM1 sequence and qualitative results are shown in Figure~\ref{fig:comp_simulation}. Our filtering pipeline (b) yields adequate estimation results of the objects given only the fingertips as inputs (red dots in the figures). Meanwhile, the (a) learning-based method~\cite{liu2021semi} would have huge errors in object pose estimation. 

We also conduct quantitative experiments in Table~\ref{tab:ablation} where different observations are fed into the filter. $\mathcal{O}_{hand}$ use the predicted 3D positions of fingertips from as observation; $\mathcal{O}_{object}$ takes the predicted object center as observations; $\mathcal{O}_{hand+object}$ uses both the aforementioned hand and object information as an observation. The results show that with more observation available, the tracking results tend to be more accurate. The tracking method is much faster than the optimization pipeline; however, the accuracy decreases since less information is provided to it.

\begin{figure}
\centering
\begin{tabular}{@{}c@{\hspace{0.3mm}}c@{\hspace{0.3mm}}c@{\hspace{0.3mm}}c@{}}
    \includegraphics[width=0.25\linewidth]{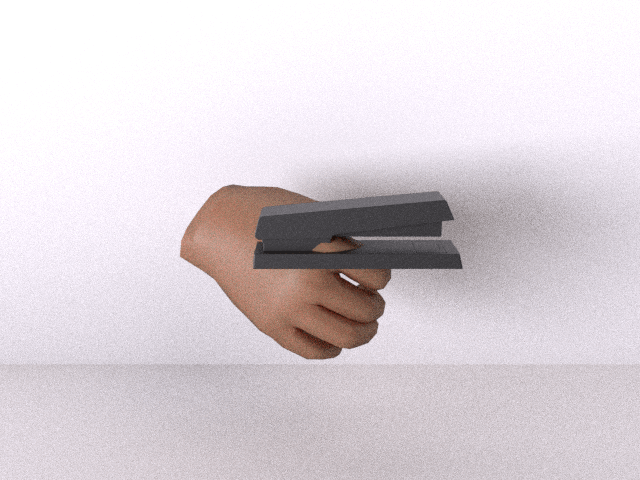} &
    \includegraphics[width=0.25\linewidth]{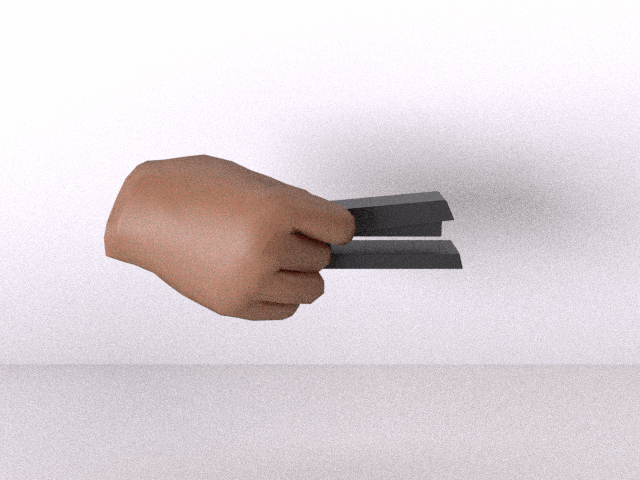}  &
    \includegraphics[width=0.25\linewidth]{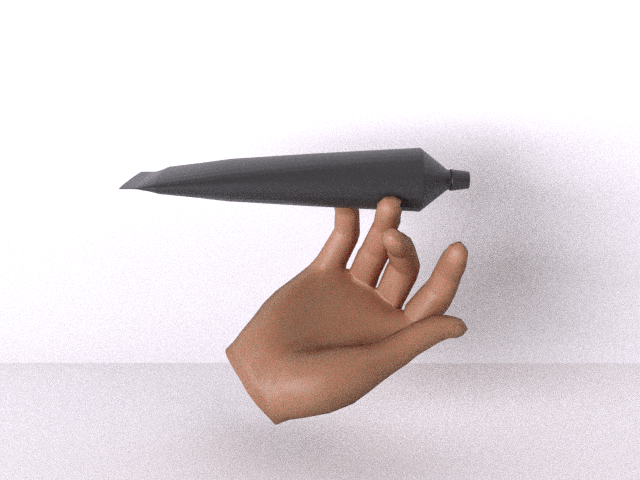} &
    \includegraphics[width=0.25\linewidth]{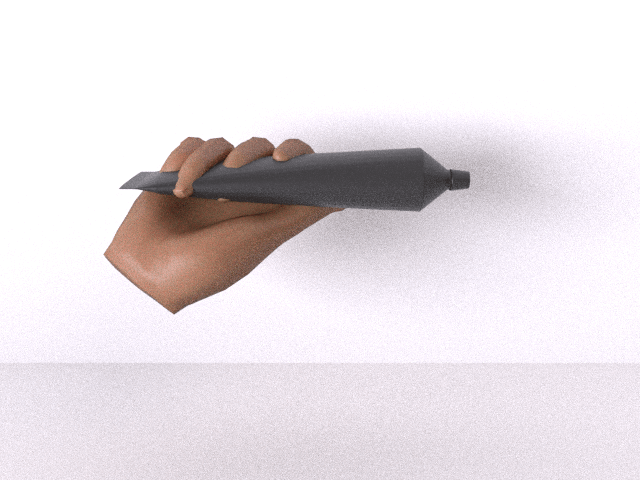}  \\
    \small (a) Initialized & \small (b) Optimized  & \small (c) Initialized & \small (d) Optimized \\
\end{tabular}
\vspace{-1mm}
\caption{\textbf{Qualitative results of contact-based pose refinement.} Given randomized initial poses (a,c) and desired contact regions, our method can refine poses using a differentiable contact model \& gradient-based optimization (b,d).  }
\vspace{-2em}
\label{fig:comp_contact}
\end{figure}

\subsection{Contact Refinement}
\label{sec:exp_contact}
The physics module can output contact information during prediction. Given the pose of the hand and the object, our GPU-based differentiable Warp kernel~\cite{warp2022} can output the signed distance function $s$ and the normal direction $\nn$ for each hand vertex $\{\nn_i, s_i\}_{i<n_{hand}}=contact(\theta, \rr, \tt)$. Following ContactOpt~\cite{grady2021contactopt}, we define a contact value $CO_i= \min(\frac{1}{s_i},1)$ for each vertex. When a vertex is within 1mm ($s_i<1$) of the object, $CO_i=1$ then means this point is in-contact. 

Given randomly initialized hand poses, our method refer to the ground truth contact information provided by ContactPose datasets~\cite{brahmbhatt2020contactpose} as the target. Since our $contact(\cdot)$ module is also differentiable, we can then optimize the poses to match the target contact status $\{CO_i, \nn_i\}$ with gradient-based optimization. 


Figure~\ref{fig:comp_contact} shows examples of the optimization process. (a) and (c) are the input randomized hand, (b) and (d) are the optimized hand-object interaction. The results are similar to how people handle those objects in daily life.

\subsection{Generalization to Human}
\label{sec:exp_human}
\begin{figure}
\centering
\begin{tabular}{@{}c@{\hspace{1mm}}c@{\hspace{1mm}}c@{}}
\includegraphics[width=0.33\linewidth,height=0\linewidth]{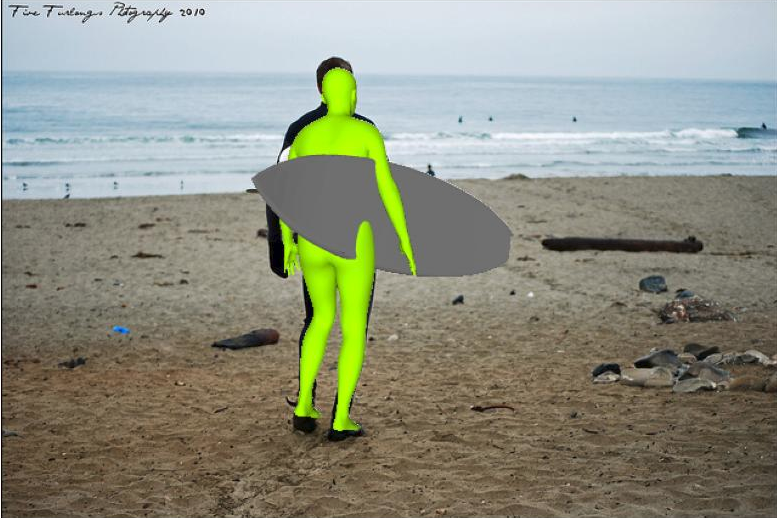} &    \includegraphics[width=0.33\linewidth,height=0\linewidth]{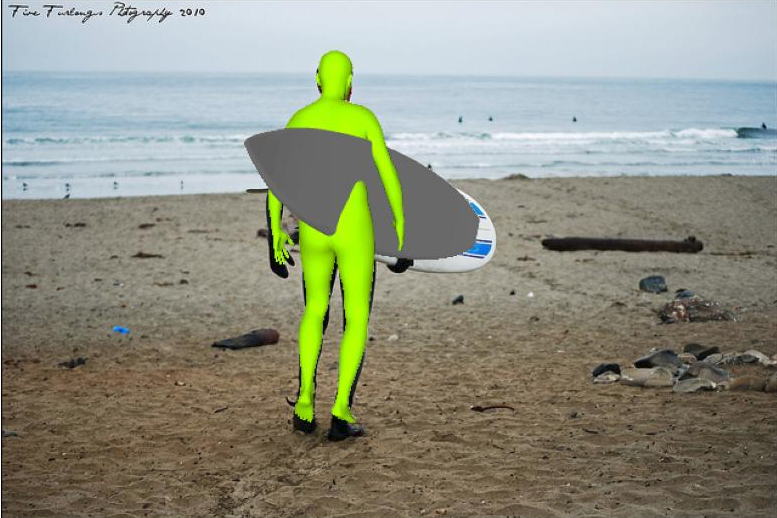} &
\includegraphics[width=0.33\linewidth,height=0\linewidth]{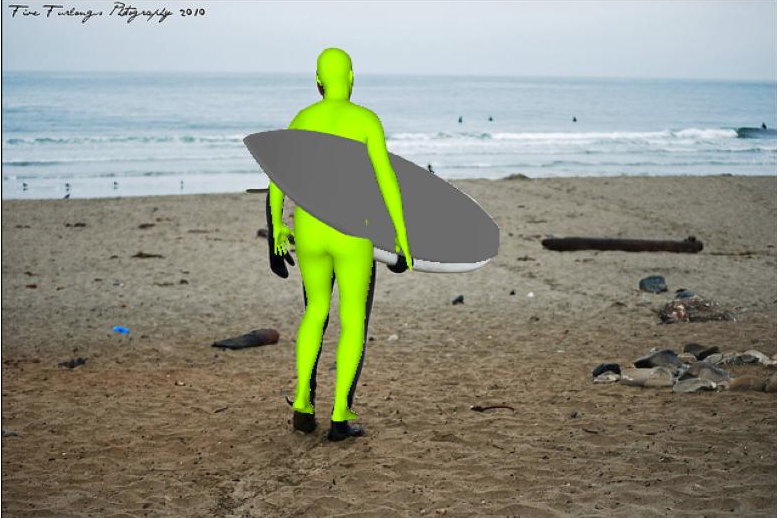}  \\
    \small (a) Initialization & \small (b) $\mathcal{L}_{render}$ & \small (c) $\mathcal{L}_{render+physics}$\\
\end{tabular}
\includegraphics[width=1\linewidth]{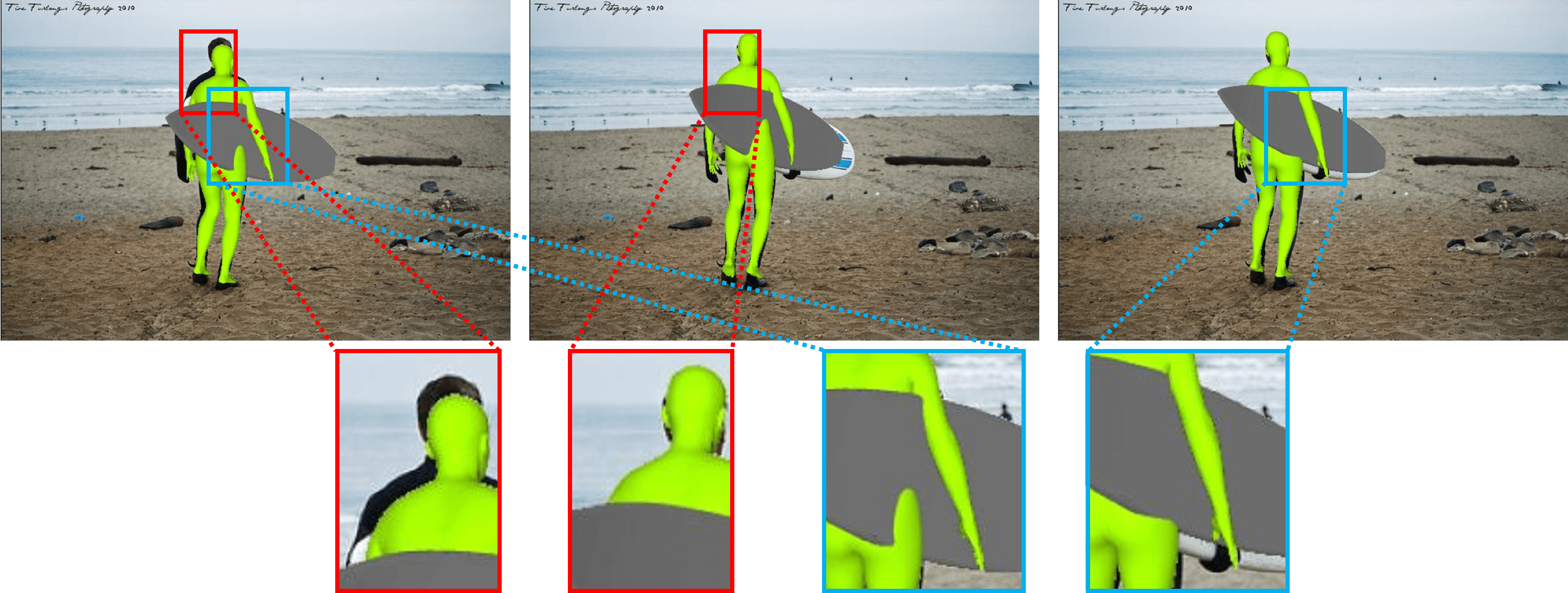}
\caption{\textbf{Human Object Pose Refinement.} Qualitative comparison between (a) initialization using~\cite{zhang2020phosa}, (b) refinement using only differentiable rendering, and (c) refinement using both differentiable rendering and physics simulator. The rendering loss helps to align the image, and the physics loss can mitigate penetrations.}
\label{fig:human_object}
\vspace{-1.75em}
\end{figure}

Our proposed differentiable priors can also be applied to human object pose estimation. With RGB images from the COCO2017 dataset~\cite{Lin_coco_2014}, we use~\cite{zhang2020phosa} to initialize both human and object poses, and then use~\cite{Kirillov_2020_pointrend} to predict the instance segmentation mask for both the human and objects. Subsequently, we replaced the MANO hand model~\cite{MANO:SIGGRAPHASIA:2017} in our pipeline by SMPL human body model~\cite{SMPL:2015} which describes the human body also by pose $\theta\in \mathbb{R}^{72}$ and shape $\beta\in \mathbb{R}^{10}$ parameters. Following the same set up as hand-object optimization, we refine human parameters $(\theta, \beta, T_{human})$ and object parameters$(\rr, \tt)$ through our pipeline using both differentiable rendering and physics priors. Our experiment shows that our method can generate accurate human-object pose estimation. As ground truth poses for human and object are unavailable, we report the intersection of union(IoU) and the interpenetration volume between human and object in Table~\ref{tab:human}. Figure~\ref{fig:human_object} shows the qualitative result of our method and the contribution of each component of the pipeline. The rendering loss in (b) helps to better align the 3D models with the RGB images but there are still non-physical collisions. The physics loss in (c) finetunes the poses and reduces such penetrations.
\begin{table}[]
    \centering
    \caption{\textbf{Quantitative evaluation on human object pose estimation} Our method can achieve higher 2D IoU and lower collision error ($cm^3$) compared to PHOSA~\cite{zhang2020phosa}.}
    \begin{tabular}{@{}lccc@{}}
    \toprule
    Method      & \multicolumn{1}{l}{human IoU} & \multicolumn{1}{l}{object IoU} & \multicolumn{1}{l}{Collision} \\
    \midrule
    PHOSA & 0.45   & 0.51  & 6818.18  \\
    +$\mathcal{L}_{render}$  & 0.64  & 0.59  & 8742.63 \\
    +$\mathcal{L}_{physics}$  & \textbf{0.68}  &  \textbf{0.64} & \textbf{666.67} \\
    \bottomrule
    \end{tabular}
    \vspace{-1.5em}
    \label{tab:human}
\end{table}

%% file: tex/conclusion.tex
In this paper, we present \methodName, a framework of differentiable rendering and physics priors for pose estimation in hand-object-interaction scenes. Two alternatives are presented to match different use cases: the optimization-based pose refinement which focuses on the prediction accuracy, and the filtering-based tracking which focuses on the runtime speed. Our proposed rendering and physics-related operators enable closed-loop optimization and tracking with self-supervision from the input RGB image sequences. Experiments show that our method yields comparable or better results in the pose estimation and contact refinement task due to our unified and general priors. We also demonstrate that our method can be applied to robotic hand manipulation and human pose estimation tasks.

We also note some limitations in our implementation. First, the optimization pipeline can fail if the initialization is poor. This problem could be alleviated by adding bounding boxes or key-point related terms to the differentiable rendering module. Second, the performance of the optimization could be improved through a faster differentiable rendering module, such as Nvdiffrast~\cite{Laine2020diffrast}. Third, the current filtering pipeline based on EKF could potentially output the uncertainty of the prediction. It would be promising to incorporate our pipeline with probabilistic prediction.